# Design a New Pulling Gear for the Automated Pant Bottom Hem Sewing Machine

[1]Ray Wai Man Kong

[1]Adjunct Professor, City University of Hong Kong, Hong Kong
[1]Modernization Director, Eagle Nice（International）Holding Ltd, Hong Kong

[2]Theodore Ho Tin Kong

[2]Graduated Student, Master of Science in Aeronautical Engineering, Hong Kong University of Science and Technology, Hong Kong
[2]Thermal-acoustic (Mechanical) Design Engineer at Intel Corporation in Toronto, Canada

[3]MIAO YI

[3]Undergraduate student, Majoring in Intelligent Manufacturing, City University of Hong Kong, Hong Kong

[3]Zerui Zhang

[4]Graduate student, Mechanical and Electrical Engineering College, Guangdong Baiyun University, China
[4]Mechanical structural engineer, Eagle Nice Ltd, Shantou, China

[1]Dr.RayKong@cityu.edu.hk, [2]theodorekong@ieee.org, [3]yimiao5-c@my.cityu.edu.hk, [4] zhang83604748@foxmail.com

**Abstract:**

Automated machinery design for garment manufacturing is essential for improving productivity, consistency, and quality. This paper focuses on the development of new pulling gear for automated pant bottom hem sewing machines. Traditionally, these machines require manual intervention to guide the bottom hem sewing process, which often leads to inconsistent stitch quality and alignment. While twin-needle sewing machines can create twin lines for the bottom hem, they typically lack sufficient pulling force to adequately handle the fabric of the pants' bottom hem. The innovative design of the pulling gear aims to address this issue by providing the necessary pulling force for the bottom hem of eyelet pants.

The research and design discussed in this article seek to solve technical challenges, eliminate the need for skilled manual operators, and enhance overall productivity. This improvement ensures smooth and precise feeding of fabric pieces in the automated twin needle sewing machine, ultimately improving the consistency and quality of the stitching.

By integrating this innovation, garment manufacturers can boost productivity, reduce reliance on manual skilful labour, and optimize the output of the production process, thereby reaping the benefits of automation in the garment manufacturing industry.

**Keywords:**

**Automation, Stitching Technology, Pull Gear, Force Analysis, Garment, Manufacturing,**

## 1      Introduction

In the rapidly evolving landscape of garment manufacturing, the integration of automation technologies has become imperative for enhancing productivity, consistency, and quality. The design of automated machinery for sewing applications is a critical component in this transformation. Under the guidance of Prof. Kong Wai Man's lean methodology for garment modernization (Kong, Kong and Huang 2024) [1], the focus is on streamlining processes, minimizing waste, and maximizing efficiency. This methodology emphasizes the importance of designing machinery that not only automates repetitive tasks but also aligns with the principles of lean manufacturing to create a more agile and responsive production environment.

Prof. Kong Wai Man's approach advocates for a holistic view of the manufacturing process, where each element of the production line is optimized to contribute to the overall efficiency. In the context of sewing machine automation, this involves a meticulous design process that considers the integration of advanced technologies such as robotics, computer vision, and IoT (Internet of Things) to create intelligent and adaptable machinery. The goal is to develop systems that can seamlessly handle various garment styles and fabrics while ensuring high precision and reducing the dependency on manual labour.

Key to this design philosophy is the incorporation of modularity and flexibility, allowing for quick adjustments and scalability in response to changing market demands. By focusing on lean principles, the machinery is designed to eliminate non-value-added activities, reduce cycle times, and enhance the flow of materials and information throughout the production process. This results in a more efficient, cost-effective, and sustainable manufacturing operation.

The design of automated sewing machinery under Prof. Kong Wai Man's lean methodology for garment modernization involves creating intelligent, adaptable, and efficient systems that align with lean principles. This approach not only improves productivity and quality but also ensures that the manufacturing process can meet the dynamic needs of the garment industry.

In the ongoing quest to enhance garment manufacturing through automation, a significant innovation has been introduced: a new design for the pulling Gear, specifically tailored for twin needles sewing machines. This advancement is particularly critical for the efficient and precise sewing of trousers, a common and intricate operation in garment production.

The double-needle sewing machine, a staple in garment manufacturing, is renowned for its ability to produce strong, parallel seams that are essential for the durability and aesthetic appeal of trousers. However, the traditional setup often requires manual intervention to guide the fabric cut pieces smoothly, ensuring consistent stitch quality and alignment. The introduction of an automated pulling Gear and control device for the design of a new automated double-needle trousers machine aims to eliminate manual skilful operators sewing the pant bottom hem by double sewing dependency, thereby streamlining the operation and enhancing overall productivity.

## 2. Literature Review

### 2.1 Garment Double Sewing Machine

The advancement of automated sewing machines has significantly impacted garment manufacturing by enhancing production speed and precision. The Garment Double Sewing Machine has emerged as a critical tool in industrial applications, particularly for high-demand production lines requiring durability and consistency in stitching (Nguyen & Lee, 2021) [2]. These machines, which operate with dual needle mechanisms, allow for simultaneous sewing of parallel stitches, resulting in more robust seam integrity and time efficiency compared to single-needle machines Studies show that double sewing mechanisms are especially beneficial in industries where uniformity and strength are essential, such as sportswear manufacturing (Kim et al., 2023) [3].

### 2.2 Pull Gear Mechanism

Integral to the functionality of advanced sewing machines is the Pull Gear Mechanism. This system, often located adjacent to the needle area, precisely manages fabric feed, ensuring even tension and consistency in stitch length (Owen & Kim, 2020) [4]. According to recent research, the pull gear plays a crucial role in controlling fabric slippage, a common issue in high-speed sewing that can lead to inaccuracies and quality defects (Singh & Lin, 2021) [5]. Manufacturers have increasingly focused on optimizing pull gear configurations to reduce machine-induced stress on delicate fabrics, which is particularly important in the production of lightweight or stretchy materials (Chen & Thompson, 2022) [6].

### 2.3 Pull Gear Force Analysis

One emerging area of study is Pull Gear Force Analysis, which examines the forces exerted by pull gears on different fabric types to optimize feed rates without compromising fabric integrity. According to Zhao et al. (2022) [7], understanding the force dynamics is essential for balancing feed speed and stitch quality, especially when dealing with multiple fabric layers. Pull gear force analysis has proven critical for reducing wear on machine parts and enhancing overall production line longevity (Li & Tang, 2023) [8]. Studies utilizing computational modelling to simulate pull gear force distribution have suggested ways to enhance machine efficiency and reduce operator intervention, which is especially relevant in fully automated production lines (Garcia & Ito, 2023) [9].

The Design and Experimental Study of Vacuum Suction Grabbing Technology to Grasp Fabric Pieces by Prof Dr Ray Wai Man Kong et al (2024) [10] provides the testing script on grabbing fabric pieces to support how to design the test script on the automated machine.

Overall, the literature highlights that while garment double sewing machines and pull gear mechanisms have significantly advanced, further research into pull gear force dynamics could lead to more refined production methods and improved machine durability.

## 3. Requirement study of the Pulling Gear in the Automated Pant Bottom Hem Sewing Machine

3.1 Pant Bottom Hem and Sewing Process

The pant bottom hem is a crucial finishing detail in garment manufacturing, particularly for trousers. This process involves folding the fabric at the bottom of the pants to create a clean, finished edge that prevents fraying and adds to the overall aesthetic of the garment. In Figure 1, the simple pant structure is a detailed description of where the pant bottom hem is located and the sewing process.

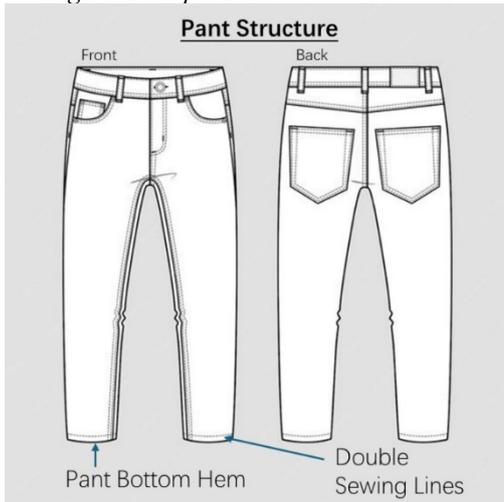

*Figure 1 Simple Pant Structure*

The bottom hem of the pants is located at the very end of each leg opening. It represents the final folded edge that encircles the lower circumference of the trousers. The width of this hem can vary based on design specifications, fashion trends, and the type of fabric utilized. Common hem widths range from narrow hems, typically measuring around 0.5 inches (1.27 cm), to wider hems that can reach up to 2 inches (5.08 cm) or more. The primary process is illustrated below:

- Design the Marker for part sewing:

The first step is to design and measure the desired length of the pant bottom opening. This involves marking the fold line where the hem will be created. Precision is key, as uneven hems can affect the garment's appearance and fit. The cutting process trims to ensure a uniform length before folding the hem.

- First Fold:

The fabric cut piece is folded up once to the inside of the pant bottom opening, usually by about 0.5 inches (1.27 cm). This initial fold helps to conceal the raw edge of the fabric.

- Second Fold:

The fabric cut piece is then folded up again, this time to the desired hem width. This second fold creates a double layer of fabric, which adds durability and a clean finish to the hem. To secure the folds, the manual work for hem sewing is required to pin or bast in place. This helps to prevent shifting and maintains the alignment of the fabric during sewing. The Automated Pant Bottom Hem Sewing Machine has a fixture to make curves of hem during the stitching process. The manual pinning and basting are not required as eliminated the operation to shorten the standard time as referred to in the future state of value stream mapping (VSM) of Lean Methodology of Garment Modernization [1].

- Machine Stitching:

The hem is typically sewn using the new design of the Automated Pant Bottom Hem Sewing Machine. A straight stitch or a blind hem stitch can be used, depending on the desired finish. The stitching is done close to the upper edge of the hem to secure the folded fabric layers together. The new design of the Automated Pant Bottom Hem Sewing Machine uses the double-needle sewing machine technology to add strength and a professional finish. The machine uses two parallel needles to create two rows of stitches, enhancing the durability of the hem and providing a decorative touch.

- After Sew Steps:

Pressing, finishing, thread trimming, ironing and packing operations are performed these steps after the new automated machine. The mid-ironing operation can be eliminated by the pulling Gear to press the hem in the sewing process.

In modern garment manufacturing, automation plays a significant role in the hemming process. The Automated Pant Bottom Hem Sewing Machine equipped with specialized features of machinery and digitalization control device, such as the newly designed pulling Gear, can significantly enhance the efficiency and consistency of hemming and sewing. The pulling Gear of the machine can precisely control the fabric feed, maintain consistent tension, and ensure uniform stitching, all of which contribute to a high-quality finished product.

By integrating advanced technologies and automation, manufacturers can achieve greater precision, reduce manual labour, and increase production speed, ultimately leading to higher productivity and better-quality garments.

### 3.2 Problem of Pulling-Gear of Sewing Process

Referring to the Lean Methodology to Garment Modernization from Prof. Ray WM Kong [1]. The requirement study is required to design the capability of the sewing machine to work all of the pants' bottom hem sewing.

The eyelet locates the equal distance between the double sew line and the edge of the hem open of the pants' bottom as shown in the eyelet of the bottom of the pants in Figure 2. The maximum eyelet diameter is related to the common design of 13mm after the requirement study. The design of the new sewing machine should not crash to the eyelet of the bottom of the pants.

Furthermore, the design of the bottom hem of the pants requires the sewing of parallel double sewing lines. No machine can perform the double sewing line to prevent the eyelet on the hem till October 2024. The sewing machine in the market can provide the simple hem of bottom pants for sewing a single line without preventing the eyelet. The eyelet design is a favour for pants for sportswear and a new trend of innovative design in the market.

*Figure 2 Eyelet in the bottom of pants*

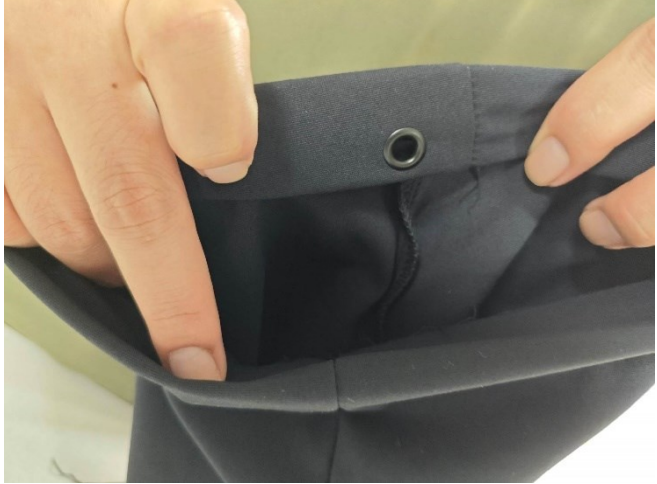

The problem of the double-needle sewing machine cannot provide sufficient pull force to pull the pants hem from the feed dog part of the sewing machine automatically. The operator adjusts the alignment of the pants' bottom parts in the double-line sewing process.

Hence, the design of the new automated machine is required to pull the fabric with the hem folding together. It is required to overcome the difficulty of technical insufficient pull force in the sewing operation. The construction of the single-needle sewing machine in Figure 3 from Schazjmd of Wikipedia [11] can be explored with more additional features of the preliminary design of the sewing machine to explore more automation ideas,

*Figure 3 The construction of the single-needle sewing machine*

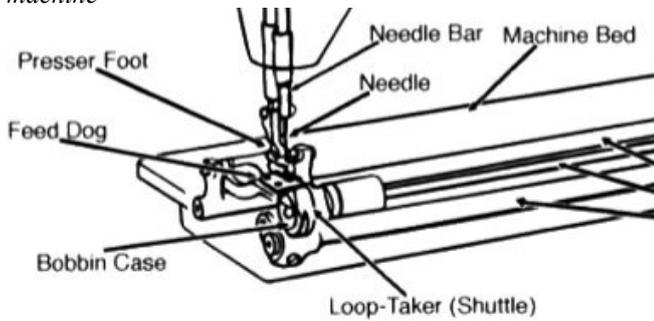

## 4. Design a New Pulling Gear
Referring to the Lean Methodology for Garment Modernization, the exploration of automated machinery design, testing and development to improve the productivity and efficiency of a particular sewing process.

The design of the pulling gear integrated into the double-needle (twin needles) sewing machine became the most important part of automation development. The design of the pulling gear is required to provide sufficient force to pull the fabric piece in the double-needle sewing machine. The required the rotatory speed and angle of the pulling Gear should align with the double needles sewing machine.

Referring to the design handbook from G. M. Maitra [12], the force between mating gears develops normally to the contacting fabric surfaces in the automated sewing machine as referred to the Figure 4.

*Figure 4 Trench Structure of Pulling Gear*

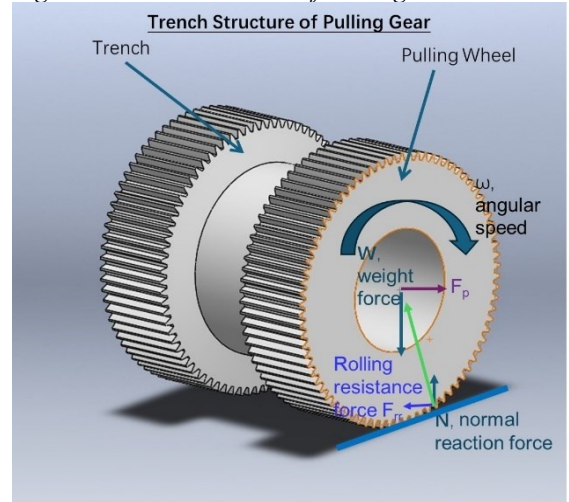

There will be a pulling force $F_p$ acting in the gear hub, which will force the gear to rotate. Due to asymmetric force distribution in the contact patch, there is going to be a normal force $N$ acting on the fabric.

The force $N$ is the vertical component of a resultant force going through the gear centre of rotation. The horizontal component of that resultant force, acting in the contact patch, which is trying to slow down the gear is exactly the rolling resistance force $F_{rr}$. The distance, $a$ is the distance of touching the fabric surface to the gear.

Since the gear is in equilibrium, the sum of forces on the x-axis, the sum of forces on the y-axis and the sum of torques acting around the centre of the gears are all zero.

X-axis forces equilibrium:
$\Sigma F_x = 0$ (1)
$F_p - F_{rr} = 0$ (2)
$F_p = F_{rr}$ (3)

Y-axis forces equilibrium:
$\Sigma F_y = 0$ (4)
$N - W = 0$ (5)
$N = W$ (6)

Torques equilibrium
$\Sigma T = 0$ (7)
$F_{rr} \cdot r_w - N \cdot a = 0$ (8)

Where $r_w$ is the radius of gear,

Replacing $N$ from (6) in (8) and solving for $F_{rr}$ gives:
$F_{rr} = (a/r_w) \cdot W$ (9)

The ratio between the distance *a* and gear radius $r_w$ is the rolling resistance coefficient *f*.

$$f = a/r_w \quad (10)$$

Replacing (10) in equation (9) gives the general formula of the rolling resistance force for a flat fabric surface.

$$F_{rr} = f \cdot W = f \cdot m \cdot g \quad (11)$$

where *m* is the pulling gear and $g = 9.81$ m/s2 is the gravitational acceleration.

The total rolling resistance force of the pulling gear is calculated and the entire gear mass is used in the equation (11). The pulling gear is rolling on a surface of fabric with the gradient α, then the formula for rolling resistance becomes:

$$F_{rr} = f \cdot W \cdot cos(\alpha) = f \cdot m \cdot g \cdot cos(\alpha) \quad (12)$$

The gradient α degree is 0 on the flat fabric surface. The *cos(0)* is 1. Equation (11) is fully applied to the search and development of the pulling force for a zero-degree gradient.

The rolling resistance coefficient of a fabric surface depends on the fabric's texture construction, materials and sewing speed. In general, for low sewing speeds, the value of the rolling resistance coefficient is constant in car tyres referred to as the rolling resistance in the Engineering Toolbox [13].

*Figure 5 Rolling resistance coefficient of car tyres*

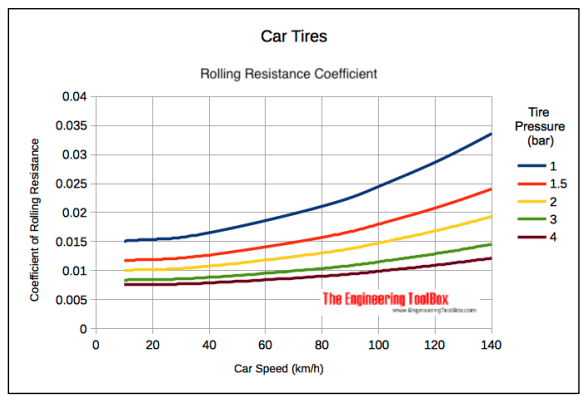

Hence, the rolling resistance coefficient is a constant when the slow sewing speed of the sewing machine according to the study from the reference [12].

The Spur Gear's transmission force $F_n$, which is normal to the tooth surface, as Spur Gear Schematic Diagram in Figure 6, can be resolved into a tangential component, $F_t$, and a radial component, $F_r$.

$$F_t = F_n \, cos\alpha_1 \quad (13)$$
$$F_r = F_n \, sin\alpha_1 \quad (14)$$

There will be no axial force, $F_x$ in the structure of spur gear. The direction of the forces acting on the gears to have the equations for tangential (circumferential) force $F_t$, axial (thrust) force $F_x$, and radial force $F_r$ about the transmission force $F_n$ acting upon the central part of the tooth flank as shown in Figure 6.

*Figure 6 Spur Gear Schematic Diagram*

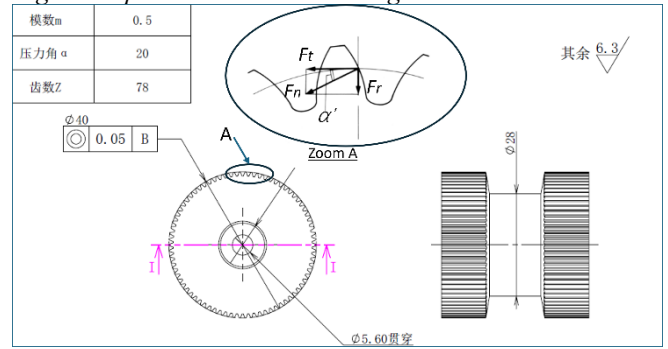

Referring to the lean methodology for Garment Modernization from Ray WM Kong [1], the design of the mechanism in the new automated machine should be required to study the requirements of the machine and motion. A pair of spur gears can grab the fabric piece by the radial force of spur gears in the drawing of the design of spur gear for the twin sewing machine Figure [7].

In the design of a pair of spur gears, the driving gear can drive the moving $F_{t1}$ tangential force on the surface of the fabric piece. The driven gear is designed to the radial force, $F_{r2}$ to keep the stable of fabric piece without changing moving direction during the sewing process.

*Figure 7 Drawing of the design of spur gear for the twin sewing machine*

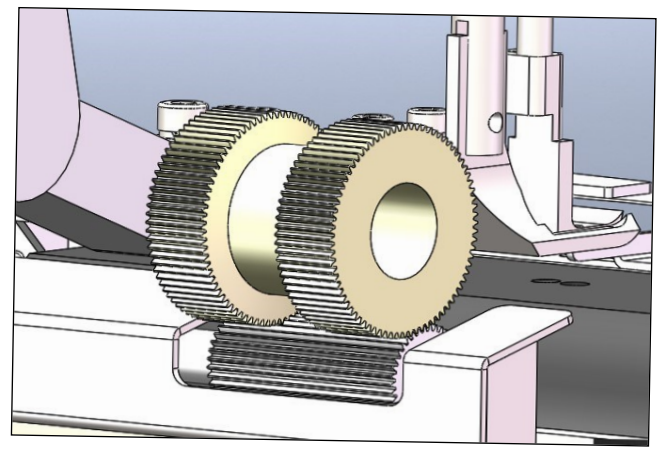

In Figure 8, the schematic force diagram in the pair of gears drives the moving $F_{t1}$ tangential force on the surface of the fabric piece. The tangential component of the drive gear, $F_{t1}$ is equal to the driven gear's tangential component, $F_{t2}$ plus the rolling resistance force from fabric piece $F_{rr}$, but the directions are opposite. The radial tangential forces, $F_{r1}$ and $F_{r2}$ keep the stable fabric piece in the moving direction during the sewing process.

*Figure 8 Schematic force diagram in the pair of gears*

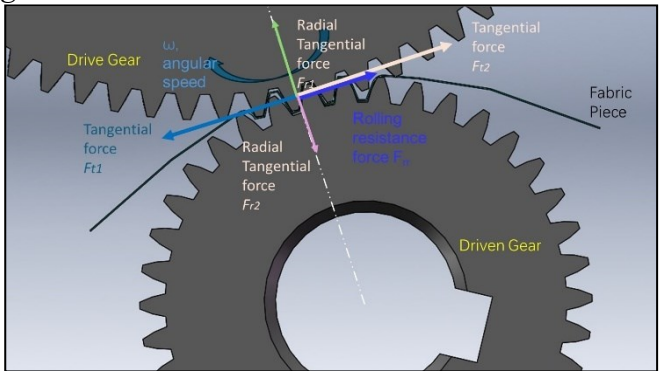

In Figure 8, the T represents the input torque from the driving gear in the formula (15).

$$T = \frac{d}{2} W_t \quad (15)$$

where $T$ is the input torque of the pulling gear motor, $W_t$ is the force of the pulling gear, d is the reference diameter of the pulling gear.

$$W_t = \frac{2T}{d} \quad (16)$$
$$W_t = F_{t1}$$

The applied force in the pulling gear comes from the pulling motor so the moving $F_{t1}$ tangential force is the same $W_t$ on the surface of the fabric piece.

$$F_{t1} = \frac{2T}{d} \quad (17)$$
$$F_{t1} = \frac{2\ (2.2\ Nm)}{40mm}$$
$$F_{t1} = \frac{2000\ (2.2\ Nm)}{40\ m}$$
$$F_{t1} = 110N$$

The force of the pulling gear is 110N. The total rolling resistance force $F_{rr}$ and tangential force $F_{t2}$ from the driven gear and fabric surface is less than the tangential force from the pulling gear $F_{t1}$.

For the case study and experiment, the fabric piece grasped the pulling gear and used the pulling gauge to pull the fabric piece. In Figure 9, the measurement of the pulling experiment at the start of the experiment, the pulling gauge showed the zero Newton (0N) force. Within 8 seconds, the fabric pieces can be pulled by a pulling gauge, so the proportional force is increased to 12N at a normal sewing speed and production sewing rate in Figure 10.

*Figure 9 The measurement of the pulling experiment at the beginning stage*

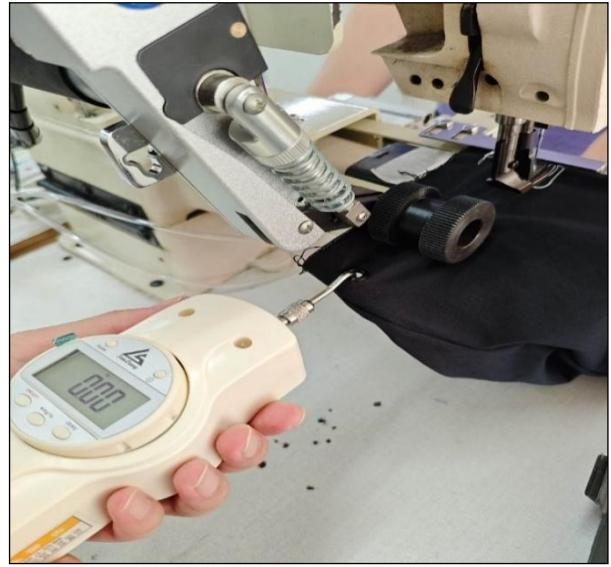

The experiment showed the required pull force to summarize the resistance force and the required tangential force of driving spur gear.

$$E(\ F_{rr}\ + F_{t2}\ ) = 12.47N$$

where E is the event of the experiment in the specific fabric piece and facility setup.

The experiment showed the applied moving $F_{t1}$ tangential force, 110N from the pulling gear motor is greater than the required rolling resistance force and required tangential force of driving spur gear in formula (18), and the pulling gear can drive the sewing process.

$$F_{t1} > E(\ F_{rr}\ + F_{t2}\ ) \quad (18)$$

*Figure 10 The peak pull force from the pulling experiment*

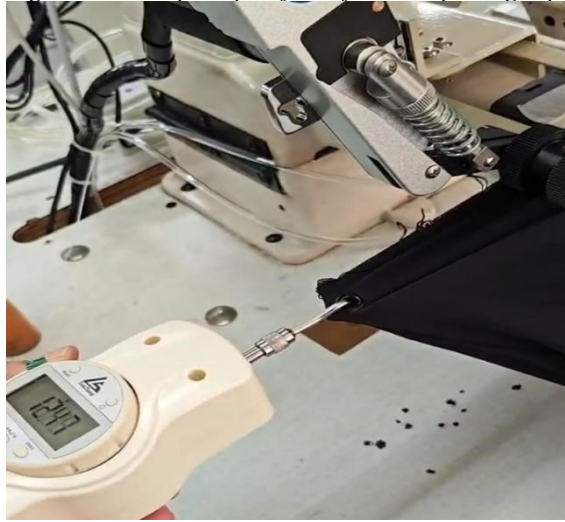

Furthermore, the various fabric pieces make an impact on the resistance force $F_{rr}$. The force analysis in the related experiment can determine the pulling force from the driving spur gear on various fabric materials and its related resistance force.

## 5. Force Analysis

In the experiment, measuring the pulling force is essential to designing the supply force of the pulling gear and mechanism. The chart below shows the progress of pulling the hem fabric piece by pull gauge. The chart in Figure [12] illustrates the progression of the pulling force over time as applied to the hem fabric piece, measured using a pull gauge. Although the data points show a generally linear trend, some fluctuations reflect variations in the applied force.

*Figure 11 The measurement of pull force-time diagram*

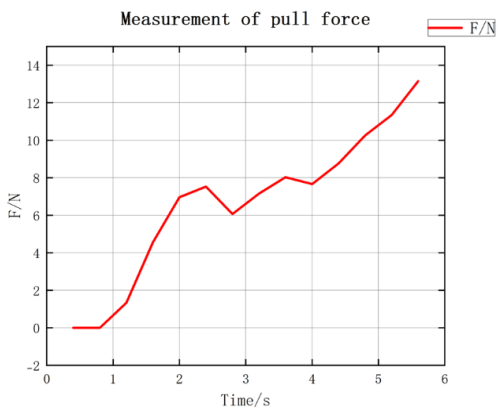

Due to the fabric being pulled manually rather than by a machine, there is some variance in the force data over time. To accurately capture these fluctuations, we extracted 20 equally spaced images from a video recording of the experiment. Each image represents a specific moment in the pulling process, allowing us to capture changes in force overtime in detail.

The data collected from these images were then imported into Origin2024, a software known for its advanced data visualization and analysis capabilities. Origin2024 allows precise control over data presentation, enabling us to create a time-series chart that visualizes the pulling force at each recorded interval. By adjusting the time intervals between data points, we were able to generate a detailed table that shows the pulling force at each moment, providing a comprehensive view of force distribution across the experiment.

Through this table and the corresponding chart, we observed that the maximum recorded pulling force, denoted as Frr (the resistance force), reached a value of 13.16 N. This resistance force is a key factor in determining the pulling gear's requirements, as it indicates the maximum force the gear must overcome to pull the fabric smoothly. Understanding this value is essential for designing a gear mechanism that can maintain a steady feed, as the pulling force must be sufficient not only to overcome Frr but also to account for any additional resistance that might occur with different fabric materials.

In addition, different fabric materials have distinct resistance characteristics, and the pulling force needed to overcome these resistances can vary significantly. By examining the force data across various fabric types, we can assess the adaptability of the pulling gear and determine whether any adjustments are needed to accommodate fabrics with higher or lower resistance.

The experiment's findings suggest that the pulling gear mechanism must be designed with flexibility to handle variations in resistance effectively. A pulling force higher than 13.16 N ensures that the gear can cope with the highest recorded resistance while providing a buffer for fabrics that might require even greater force. This adaptability is crucial for maintaining consistency in the sewing process, as it reduces the risk of stitch irregularities and enhances the overall quality of the final product.

## 6. Conclusion

In conclusion, the design of a new pulling gear for the automated pant bottom hem sewing machine brings significant benefits to the garment manufacturing industry. By eliminating the need for manual intervention and skilled operators, this innovation streamlines the operation and enhances overall productivity. The automated pulling Gear ensures smooth and precise feeding of fabric cut pieces, resulting in consistent stitch quality and alignment. This improvement in efficiency and quality leads to increased productivity and reduced labour costs for garment manufacturers. Additionally, the integration of automation technologies in the pant bottom hem sewing machine aligns with lean principles, optimizing the flow of materials and information throughout the production process. Overall, the introduction of the new pulling gear design for the automated pant bottom hem sewing machine improves productivity, consistency, and quality, making it a valuable advancement in the garment manufacturing industry.

Below benefits of pulling gear can facilitate enhanced fabric piece control, integrated control devices, automation with efficiency, adaptability and flexibility, quality and consistency.

### 6.1 Enhanced Fabric Piece Control

The newly designed pulling Gear is engineered to provide superior control over the fabric as it moves through the sewing machine. By maintaining consistent tension and alignment, the pulling Gear ensures that the fabric feeds smoothly and accurately, resulting in uniform stitches and reducing the likelihood of errors such as puckering or misalignment.

### 6.2 Integrated Control Device

Alongside the pulling Gear, an additional control device is incorporated into the system. This device is responsible for synchronizing the pulling Gear's operation with the sewing machine's needle movements. Through precise coordination, the control device ensures that the fabric is advanced at the optimal rate, matching the sewing speed and maintaining stitch integrity.

### 6.3 Automation and Efficiency
The integration of the pulling Gear and control device transforms the double-needle sewing machine into a more automated system. This automation reduces the need for manual handling, allowing operators to focus on other tasks and ultimately increasing the throughput of the sewing operation. The result is a more efficient production line capable of handling higher volumes with consistent quality.

### 6.4 Adaptability and Flexibility
The design of the pulling Gear and its control system is modular, allowing for easy adjustments to accommodate different fabric types and thicknesses. This flexibility is crucial for garment manufacturers who deal with a variety of materials, ensuring that the machinery can be quickly adapted to meet diverse production requirements.

### 6.5 Quality and Consistency
By automating the fabric feeding process, the new pulling Gear design significantly enhances the consistency of the sewing operation. This consistency translates to higher quality finished products, with uniform seams and reduced defects, meeting the stringent standards of modern garment manufacturing.

In conclusion, the introduction of the new pulling gear design, along with its accompanying control device, represents a substantial leap forward in the automation of twin needles sewing machines for trouser operations. This innovation not only improves the efficiency and quality of the sewing process but also aligns with the broader goals of modern manufacturing to create more agile, adaptable, and high-performing production systems.

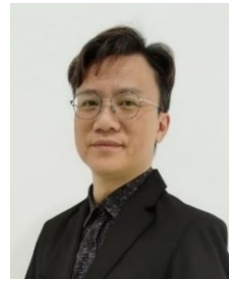

Prof Dr Ray Wai Man Kong (Senior Member, IEEE and Member IET) Hong Kong, China. He received a Bachelor of General Study degree from the Open University of Hong Kong, Hong Kong in 1995. He received an MSc degree in Automation Systems and Engineering and an Engineering Doctorate from the City University of Hong Kong, Hong Kong in 1998 and 2008 respectively.

From 2005 to 2013, he was the operations director with Automated Manufacturing Limited, Hong Kong. From 2020 to 2021, he was the Chief Operating Officer (COO) of Wah Ming Optical Manufactory Ltd, Hong Kong. He is currently a modernization director with Eagle Nice (International) Holdings Limited, Hong Kong. He holds an appointment, as an Adjunct Professor of the System Engineering Department at the City University of Hong Kong, Hong Kong. His research interest focuses on Intelligent Manufacturing, Automation, Maglev Technology, robotics, Mechanical Engineering, Electronics and System Engineering for industrial factories.

Prof. Dr. Kong Wai Man, Ray is chairman of the Intelligent Manufacturing Technology Committee of Doctor Think Tanks Academy, Hong Kong and Vice President of CityU Engineering Doctorate Society, Hong Kong. He has published 21 intellectual properties and patents in China.

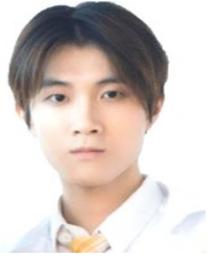

**Theodore Ho Tin Kong** (MIEAust, Engineers Australia) received his Bachelor of Engineering (Honours) in mechanical and aerospace engineering from the University of Adelaide, Australia, in 2018. He then earned a Master of Science in aeronautical engineering (mechanical) from HKUST—Hong Kong University of Science and Technology, Hong Kong, in 2019.

He began his career as a Thermal (Mechanical) Engineer at ASM Pacific Technology Limited in Hong Kong, where he worked from 2019 to 2022. Currently, he is a Thermal-Acoustic (Mechanical) Design Engineer at Intel Corporation in Toronto, Canada. His research interests include mechanical design, thermal management and heat transfer, and acoustic and flow performance optimization. He is proficient in FEA, CFD, thermal simulation, and analysis, and has experience in designing machines from module to heavy mechanical level design.

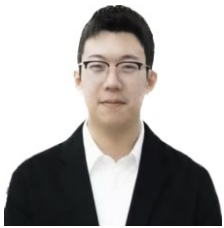

Miao Yi (Undergraduate student in Intelligent Manufacturing and Engineering at City University of Hong Kong) in 2024. He has research experience in modal electromagnetic ultrasonic transducers and machine learning applications in intelligent manufacturing.

He worked as a Research Assistant at Hangzhou Zheda Jingyi Electromechanical Technology Co., Ltd., and is currently conducting research on AI integration in manufacturing under Professor Meng Yuquan. His interests include intelligent manufacturing and sustainable engineering solutions.

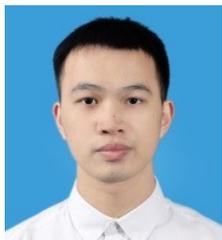

**Zerui Zhang (Graduate)** received a Bachelor of Engineering degree in Mechanical Design, Manufacturing and Automation at Mechanical and Electrical Engineering College, Guangdong Baiyun University 2019.

Before graduation, he was offered a reserve cadre position at BIEL CRYSTAL in Huizhou, Guangdong Province, where he will continue to work as a mechanical engineer until 2021. Currently, he is a mechanical structural engineer at EAGLE NICE in Shantou, Guangdong Province. He is committed to the study and research of mechanical design, mechanical simulation, machining and assembly technology. He is proficient in SOLIDWORKS, AUTO CAD etc 3D software modelling, and has the research and development experience of product parts to the whole machine.